\RequirePackage{fix-cm}
\documentclass{svjour3}                     
\smartqed  

\usepackage[table]{xcolor}
\usepackage{enumitem}

\usepackage[export]{adjustbox}
\usepackage{amsmath,latexsym,amsbsy,amssymb}
\usepackage{longtable}

\usepackage{subcaption}

\usepackage{graphicx}
\usepackage{amsmath}
\usepackage[numbers]{natbib}
\usepackage{hyperref}
\usepackage{xurl}
\usepackage{caption}
\usepackage{rotating}
\usepackage{lscape}

\journalname{Applied Intelligence}
\begin{document}

\title{Leaf Recognition Using  Convolutional Neural Networks Based  Features\thanks{The first two authors have equal distribution}
}

\titlerunning{Leaf Recognition Using  Convolutional Neural Networks Based  Features}        

\author{Boi M. Quach \and
        Dinh V. Cuong  \and
        Nhung Pham \and
        Dang Huynh \and
        Binh T. Nguyen
}



\institute{Boi M. Quach \at
              Dublin City University, Ireland  
           \and
           Dinh V. Cuong \at
              Dublin City University, Ireland
              \and
           Nhung Pham \at
             University of Science, Ho Chi Minh City, Vietnam\\
             Vietnam National University in Ho Chi Minh City
              \and
              Huynh T. Dang  \at
              AISIA Research Lab, Ho Chi Minh City, Vietnam \\
              Hong Bang International University, Ho Chi Minh City, Vietnam
              \and
              Binh T. Nguyen (Corresponding Author) \at
              AISIA Research Lab, Ho Chi Minh City, Vietnam \\
              University of Science, Ho Chi Minh City, Vietnam\\
              Vietnam National University in Ho Chi Minh City\\
              \email{ngtbinh@hcmus.edu.vn}
              }

\date{Received: date / Accepted: date}

\maketitle

\begin{abstract}
There is a warning light for the loss of plant habitats worldwide that entails concerted efforts to conserve plant biodiversity. Thus, plant species classification is of crucial importance to address this environmental challenge. In recent years, there is a considerable increase in the number of studies related to plant taxonomy. While some researchers try to improve their recognition performance using novel approaches, others concentrate on computational optimization for their framework. In addition, a few studies are diving into feature extraction to gain significantly in terms of accuracy. In this paper, we propose an effective method for the leaf recognition problem. In our proposed approach, a leaf goes through some pre-processing to extract its refined color image, vein image, xy-projection histogram, handcrafted shape, texture features, and Fourier descriptors. These attributes are then transformed into a better representation by neural network-based encoders before a support vector machine (SVM) model is utilized to classify different leaves. Overall, our approach performs a state-of-the-art result on the Flavia leaf dataset, achieving the accuracy of 99.58\% on test sets under random 10-fold cross-validation and bypassing the previous methods. We also release our codes\footnote{Scripts are available at \url{https://github.com/dinhvietcuong1996/LeafRecognition}} for contributing to the research community in the leaf classification problem. 
\end{abstract}

\section{Introduction}

There are approximately a half of million plant species worldwide, and many of them have not been recorded recently. These are sources of food, medicine, recreation, genes, poisons, animal feed, and building material. It may ultimately lead to extinction for many species within the foreseeable future. Therefore, a plant classification system is needed.
Based on flowers, fruits, leaves, and other structures, one can quickly identify plants in different types of categories. Although living plants are represented in 3D objects, still images capture 2D projections. Thus, the use of flowers and fruits to extract features may increase complexity as their attributes can be shown with great clarity in the three dimensions. At the same time, plant leaves are two-dimensional. Besides, most plant species have unique leaves that are different from each other based on many features, such as color, shape, texture, and the margin \cite{beghin2010shape}. Therefore, the classification of plants based on leaves is efficient and effective in recognizing species of plants. Furthermore, leaves are easily found and collected anytime, whereas flowers or fruits are dependent on seasons and types of plants.

Leaf recognition plays a crucial role in plant taxonomy, and its applications also help botanists rescue an enormous number of plant species. Thus, finding a leaf classification system's better performance is always an attractive topic for many researchers to try other approaches. The traditional image processing method to classify and identify plants' species uses hand-engineered features and then trains them alongside standard machine learning models like Support Vector Machines (SVMs), Random Forest, and K-Nearest Neighbors (KNNs). However, using convolutional neural networks (CNNs), deep learning approaches have recently seen a significant breakthrough in various image classification tasks considering these architectures' apparent empirical success.

For academic research purposes, many leaf databases belonging to plant species are published. In this paper, the Flavia leaf dataset \footnote{This dataset is downloadable at http://flavia.sourceforge.net/} is chosen by its popularity, although there are also articles that use other leaf databases. This dataset is initially released by \cite{wu2007leaf}, consisting of 32 classes and 1907 leaves. The current state-of-the-art results so far on the Flavia database are presented by Turkoglu and Hanbay in 2019 with the accuracy of $99.1\%$ \cite{turkoglu2019recognition} and $99.42\%$ \cite{8875911}. It is a considerable challenge to achieve the same result or even more than that. Nevertheless, our research proposed a novel method that combines handcrafted features and CNN-based features to improve performance.

The rest of the paper can be organized as follows. First, we briefly review the related works on leaf recognition based on different techniques. The following section describes the handcrafted features implemented in our proposed method. We then outline the classification model architecture. After that, we express our experimental setups along with the corresponding results. Ultimately, the paper ends with our conclusion and future works.

\section{Related Work}

There are two feature categories employed in plant identification: handcrafted features and deep learning features. For the former, a subset of shapes, colors, and texture features have been utilized in most studies considering they represent the main characteristics of a specific leaf. It is worth noting that the process of dealing with these problems, such as different scales, lighting intensity, resolution, background clutter, and appearance, is extremely time-consuming. Therefore, deep learning has been developed to reduce such issues but still helps the classification task achieve a good performance. Nevertheless, in terms of accuracy, if computer vision techniques can tackle these limitations before extracting engineered features, the performance is approximately equal to that of neural networks. Indeed, the highest performance of using purely handcrafted attributes so far is 99.43\% \cite{su2020fast} while that of deep learning techniques is 99.42\% \cite{8875911} in the Flavia dataset. Multiple studies have employed both hierarchies for this plant recognition task - conventional machine learning algorithms and neural networks. They can be implemented separately or combined to achieve better accuracy. In this literature review, we also describe previous studies according to each type of used method.

\subsection{KNN based methods}
K-Nearest Neighbor (KNN) is also a standard classifier in plant recognition. Using this technique, one compared each leaf's features in tandem to find the closest similarities, and then it could be easily recognized into a well-suited group.
In 2015, Munisami et al.\cite{munisami2015plant} used KNN to classify 32 different plants classes in Folio dataset and obtained an accuracy of 87.3\%. However, the number of pictures for each species is limited, with only 20 images for each and 640 images in total. Given the Faliva dataset, Kumar et al.\cite{Kumar2016LeafCB} extracted shape and edge attributes before putting these features into the KNN model, and then they could obtain an average precision of 94.37\%. In 2020, Su et al. \cite{su2020fast} proposed five multi-scale triangle representations for each leaf image after extracting the leaf's form to improve the classification performance. This approach used KNN as an optimization method and could achieve outstanding performance with an accuracy of 99.43\%.

\subsection{ SVM based methods}

Some other papers implemented Support Vector Machine (SVM) classifiers in the leaf classification due to their robustness and regularization. The principle of SVMs is looking for decision boundaries to separate features consistently. For example, related to the Flavia dataset, Zhang et al. \cite{zhang2012plant} applied SVMs to identify 32 different classes based on leaf shape and texture descriptors. The experimental performance gained an average accuracy of over 93.8\%. In the same year, Priya et al.\cite{Priya2012AnEL} presented a scheme that extracted shape and vein features that originated from five fundamental attributes. The final result showed that using SVM classifiers could attain an accuracy of 94.5\%  compared to the previous approach. Finally, Ahmed et al. \cite{ahmed2016automatic} implemented a system where fifteen features were fed into a process, including feature extraction, normalization, dimensionality reduction, and ultimately recognition. This specific approach could achieve an average accuracy of 87.40\% on the Flavia dataset.

Wang and colleagues \cite{wang2014plant} applied the intersecting cortical model (ICM) for extracting shape and texture features. In order to lessen a burden in the training process, principal component analysis (PCA) was used to reduce the dimensionality of sets of feature vectors that were then fed into the SVM model. In this study, the experimental results achieved an accuracy of 97.82\% with the Flavia dataset. Ghasab et al. \cite{ghasab2015feature} employed a decision-making algorithm, namely ant colony optimization (ACO), that could build a search space for optimizing paths between features. A set of features such as shape, texture, color, and morphology are discriminated on these graphs before utilizing SVM classifiers to predict the corresponding species. The performance of this research obtained an accuracy of 96.25\% on the Flavia dataset.

The further study presented by using different techniques to leaf contour and shape features, classifying plant species into 32 categories \cite{khmag2017recognition}. They used the SVM model for the Flavia dataset and achieved an accuracy of 97.7\% for the proposed method. Familiar to Uluturk's approach, Turkoglu et al.\cite{turkoglu2019recognition} conducted the division of leaf images into bisection and quarter sections. Also, by extracting five basic feature engineers such as shape, color, vein, contour, and texture via various methods, the approach obtained 99.1\% classification accuracy based on the efficiency of SVMs.

\subsection{Fusion based methods}

Apart from these models, others methods have been used or combined in leaf image classification. For instance, after conducting experiments with Support Vector Machines, k-Nearest Neighbor, Naive Bayes, and Random Forest classifier, Ali et al. \cite{caglayan2013plant} achieved a better success rate approaching 96\% with the Random Forest method when they only used shape and color features for plant recognition. 

Further, there are others works in which five standard modalities are taken into account. For example, all three models, such as SVM, Naïve Bayes, and decision trees, were applied by Eid et al. \cite{eid2015leaf} in 2015. Consequently, the corresponding experimental results were 91.8\%, 88.3\%, and 85.7\%, respectively. One of the most impressions in this study was that the authors combined a subset of features and then found that 97\% of leaf identification based on ten different biometric information extracted from shape and vein features was the best performance. 

Along with the combination of these modalities, the fusion of various techniques for leaf recognition was conducted by Mittal et al.\cite{mittal2018combined} in 2018, followed in a maximum accuracy of 93.1\% for using three classifiers Naïve Bayes, and a decision tree and SVM in tandem. Another technique was proposed in \cite{aakif2015automatic} using Artificial Neural Network (ANN) after collecting different digital morphological attributes, extracting Fourier descriptors from contours, and defining shape features. They ultimately obtained a classification accuracy of 96\% in the Flavia dataset. Besides, the Extreme Learning Machine (ELM) was also utilized for classification along with a hybrid system was proposed to extract the distinctive features of the leaf \cite{turkouglu2019plant}, which gained a recognition rate of 98.31\% for Flavia data sets.

\subsection{Deep learning-based methods}

There also have been many attempts to employ deep neural networks on leaf recognition problems, including \cite{HYBRID_CNNSVM,DUAL_MODEL,EXTERNAL_CNN,JEON_CNN,VEIN_CNN,MULTISCALE_CNN}. The classic research on the Flavia dataset was executed by Wu et al.\cite{wu2007leaf}, which applied Probabilistic Neural Network (PNN) with image processing techniques. In this study, thirty-two species were classified using 12 attributes in five principal components with a high performance of 90,312\%. In terms of feature extraction, Türkoğlu et al. \cite{8875911} developed a method using deep features which were extracted from the fc6 layer of AlexNet and VGG16 models. Then, Principal Component Analysis (PCA) was applied for dimensionality reduction of these features. Their result achieved for Flavia data sets was 99.42\% after utilizing the KNN classifier. 
Another approach \cite{HYBRID_CNNSVM} trained multiple autoencoders and deep CNN to extract features and then classified leaves by SVM classifiers. Similarly, the study \cite{DUAL_MODEL,EXTERNAL_CNN} utilizes a dual-path deep convolutional neural network to collect leaf image features. Specifically, these proposed neural network comprises two branches, one for learning features from the color leaf image and another for capturing features from the texture patch (which is the central region of the leaf). These branches later join to combine the information and are inputted to a multi-layer perceptron. Thus, the entire network is jointly trained together. One can find other references at \cite{VEIN_CNN,EXTERNAL_CNN,JEON_CNN}.

\section{Methodology}
This section presents our proposed architecture, the image processing for each leaf image, and how to construct a suitable model and features for the main problem. 

\subsection{Image Processing}
\begin{figure}[!htpb]
	\centering
	\begin{subfigure}{0.35\textwidth}
		\centering
		\includegraphics[width=\linewidth]{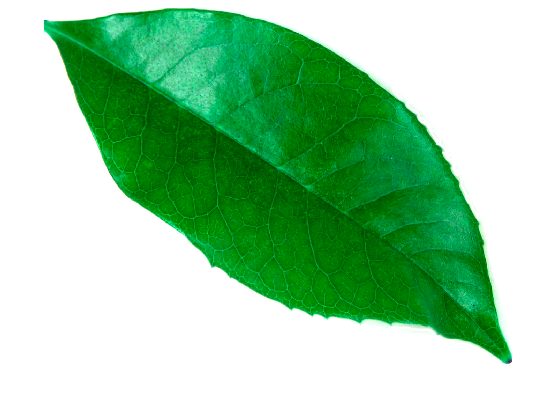}
		\caption{}
		\label{fig:f1} 
	\end{subfigure}\quad
	\begin{subfigure}{0.4\textwidth}
		\centering
		\includegraphics[width=\linewidth]{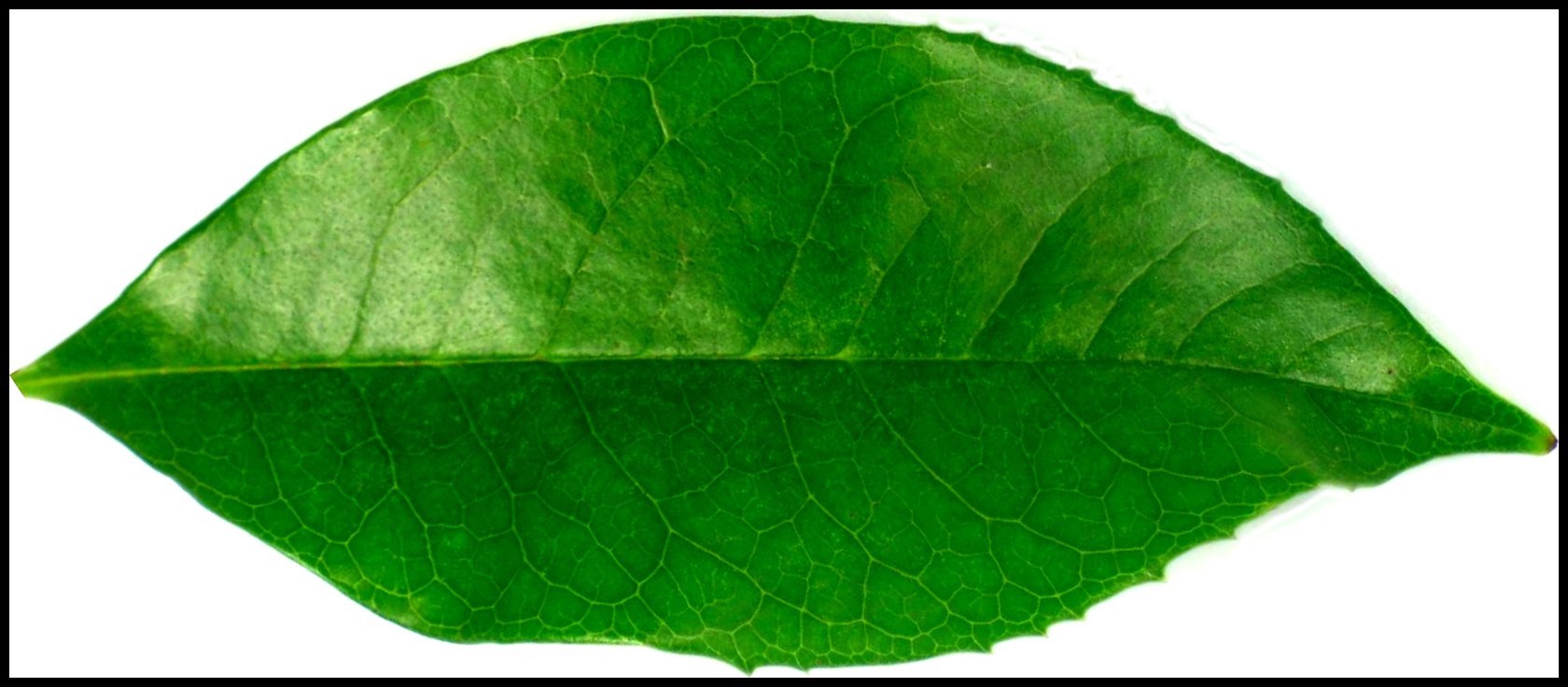}
		\caption{}
		\label{fig:f2}
	\end{subfigure}
	\caption{(a) Original image (b) Processed image }
	\label{fig: f24}
	\centering
\end{figure}

Images downloaded from the Flavia dataset are in various formats along with different resolutions and quality. The first step is image processing since dataset images need to be converted into the same format to get better feature extraction. This stage is generally detailed in Figure~\ref{fig: f24} as follows:\\
\textbf{Step 1:} Based on the direction of the main vein, leaves images are rotated and aligned.\\
  Principal Component Analysis (PCA) is applied to extract the main vein by transforming from the 2-dimensional leaf image matrix into a 1-dimensional column vector. PCA often makes the size of the corresponding covariance matrix too large and computes the eigenvectors and eigenvalues. The first eigenvector and eigenvalue will be respectively the direction and the magnitude of the main vein.\\
\textbf{Step 2:} Cropping the image to remove the unwanted background.\\
  After determining the maximum contour of the leaf image, the rectangular containing the leaf is retained and carries out the removal of the remaining image.\\
\textbf{Step 3:} Resizing the dataset image to the same size. \\
  All new input leaf images are in 1600x1600 resolution in a common format. Thus, the restriction on the direction of leaves is reduced considerably.

\begin{figure}[!htpb]
    \centering
   \begin{subfigure}{0.3\textwidth}
       \centering
       \includegraphics[width=\linewidth]{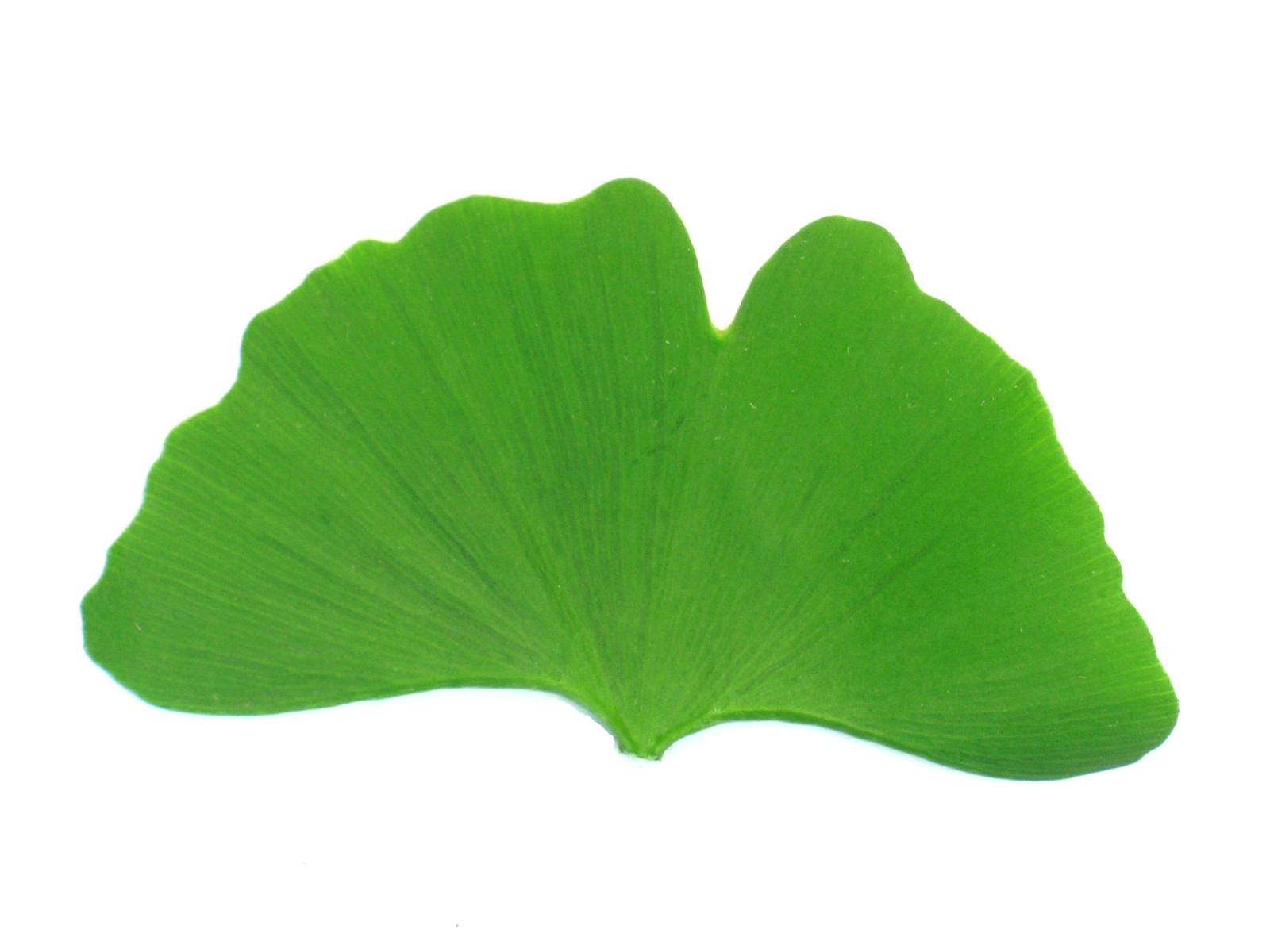}
       \caption{}
       \label{fig:f1} 
    \end{subfigure}\quad
    \begin{subfigure}{0.3\textwidth}
       \centering
       \includegraphics[width=\linewidth]{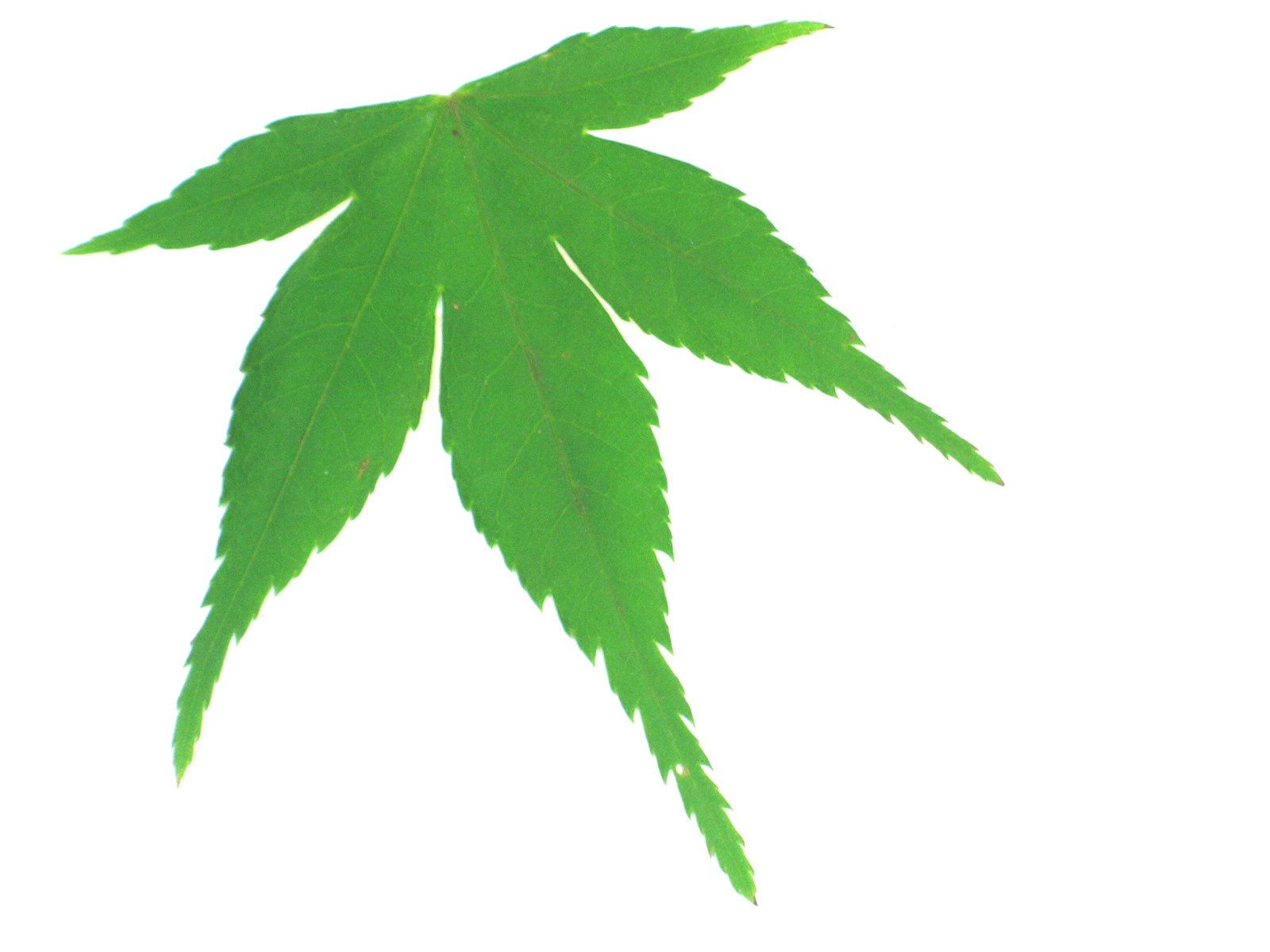}
       \caption{}
       \label{fig:f2}
    \end{subfigure}
    \begin{subfigure}{0.3\textwidth}
       \centering
       \includegraphics[width=\linewidth]{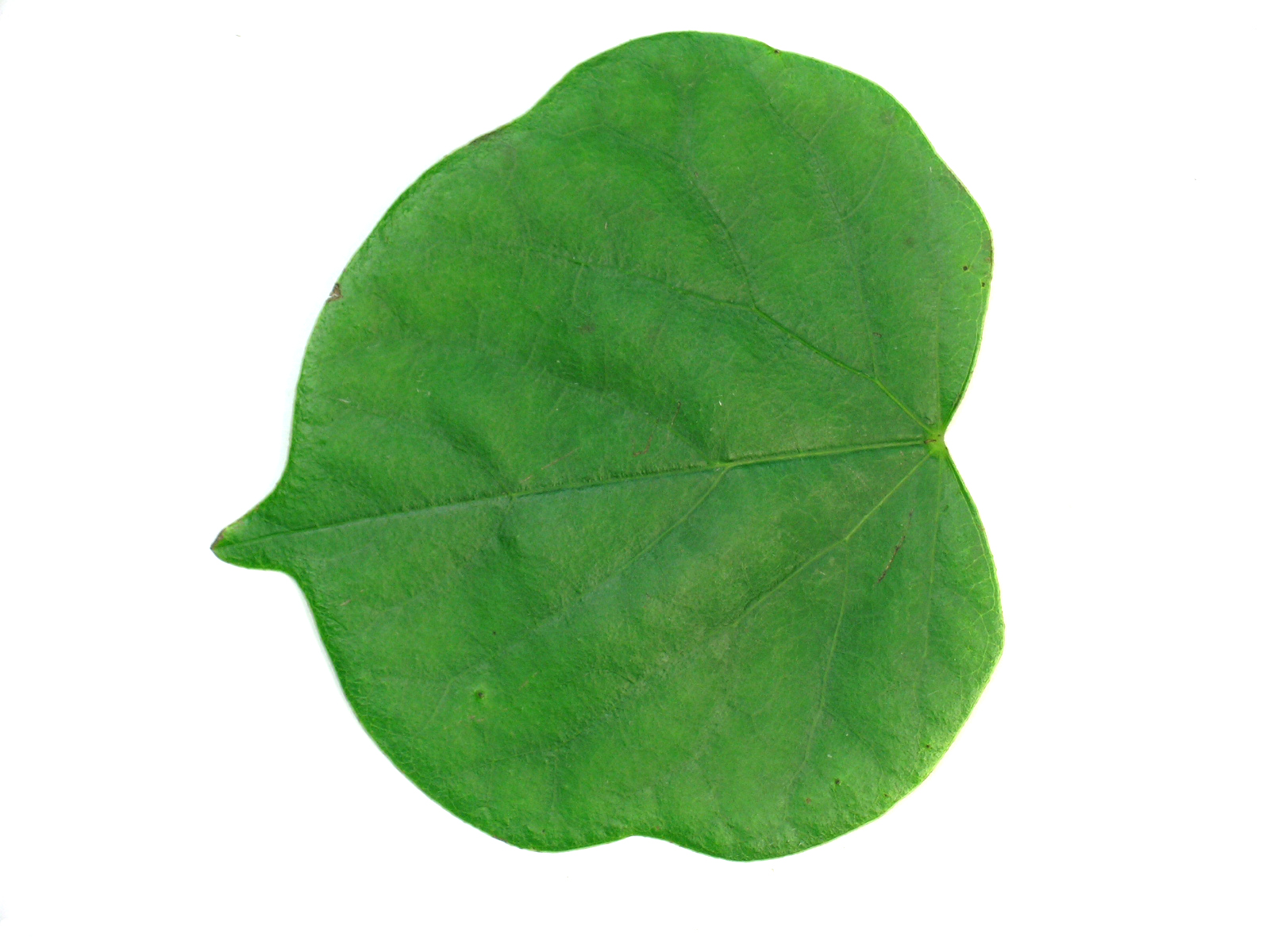}
       \caption{}
       \label{fig:f3}
    \end{subfigure}
    \caption{(a) Ginkgo and maidenhair leaf (b) Japanese maple leaf (c) Chinese redbud leaf} 
    \label{fig: figure8}
    \centering
\end{figure}

The limitation of rotation in the image processing step is that we cannot precisely detect some groups' prominent veins. For example, most leaves have a midrib (main vein), which travels the length of the leaf and branches to each side to produce veins of vascular tissue. However, in dicotyledons, the leaf's veins have a net-like appearance, forming a pattern known as reticulate venation. Ginkgo and maidenhair leaves in Figure~\ref{fig: figure8}(a) are examples of a plant with dichotomous venation. Besides, the leaves with three main veins from the base, as shown in Figure~\ref{fig: figure8}(b), is also a complicated barrier in finding which vein is the main vein to rotate the image. Another problem is for the orbicular leaves (as depicted in Figure~\ref{fig: figure8}(c)), which have a more or less circular leaf shape in which the width and length of the lamina are equal, or nearly so. In these leaves, the veins are too dense or almost disappear in the foreground.

\subsection{Handcrafted Feature}
The next step is to convert the gray image from the RGB format. Typically, the equation (\ref{eqn:eq1}) can be applied to transfer the RGB values of a pixel into the corresponding grayscale value as follows:
\begin{equation}
\label{eqn:eq1}
  gray = 0.5870G + 0.2989R + 0.1140B,
\end{equation}
where R, G, B are red, green, blue components, respectively.

\begin{figure}[!htpb]
    \centering
   \begin{subfigure}{0.3\textwidth}
       \centering
        \includegraphics[width=\linewidth]{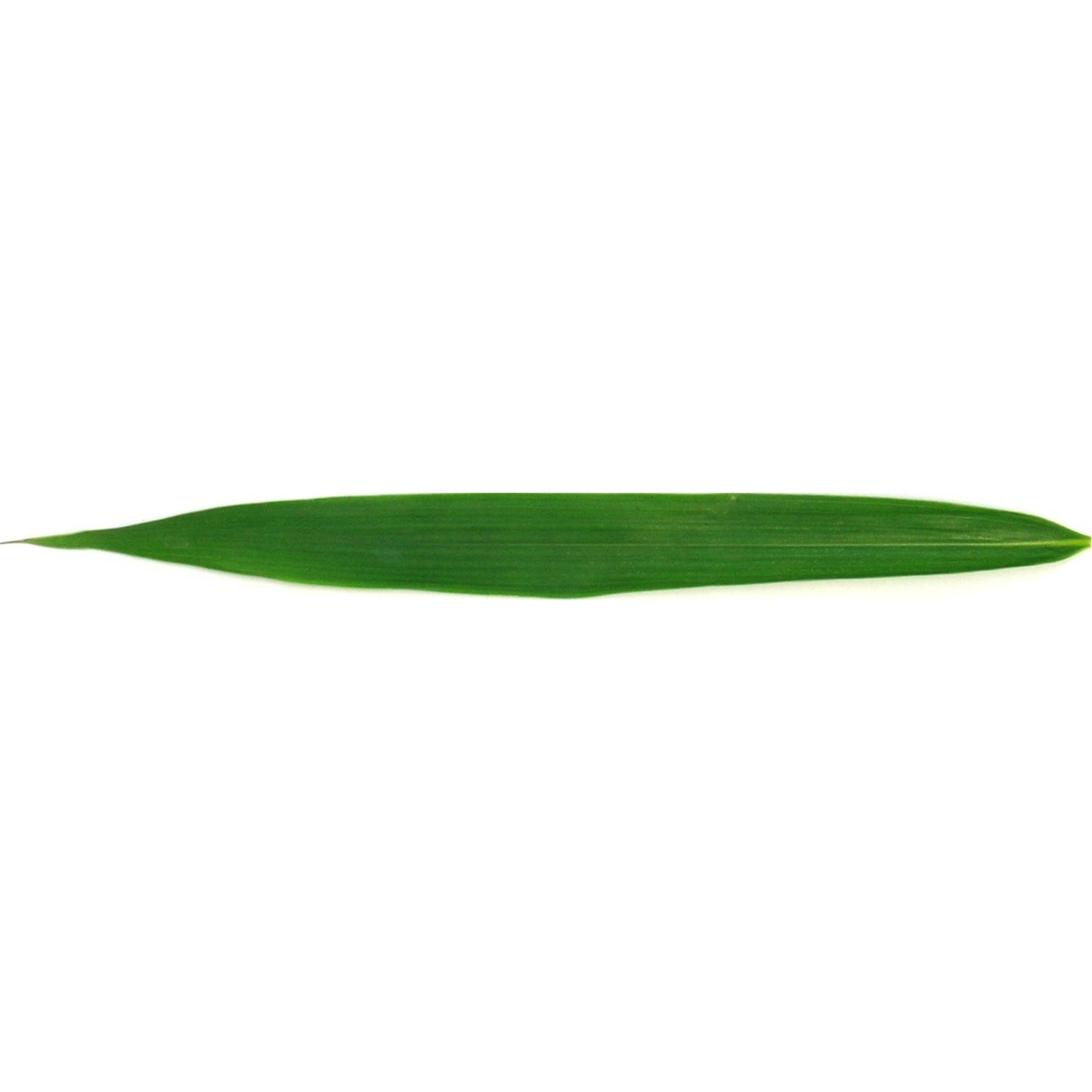}
       \caption{}
       \label{fig:f1} 
    \end{subfigure}\quad
    \begin{subfigure}{0.3\textwidth}
       \centering
        \includegraphics[width=\linewidth]{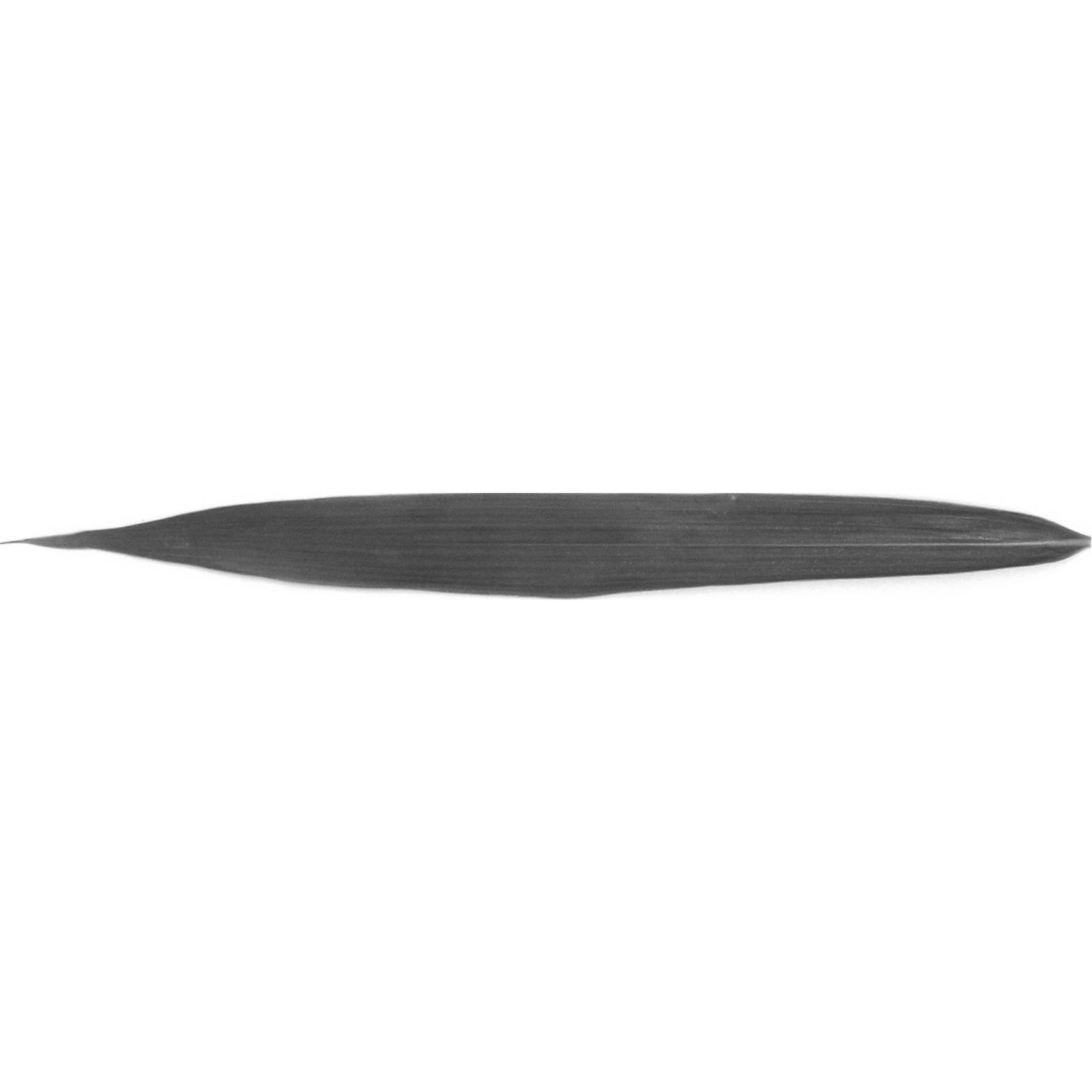}
       \caption{}
       \label{fig:f2}
    \end{subfigure}
    \begin{subfigure}{0.3\textwidth}
       \centering
        \includegraphics[width=\linewidth]{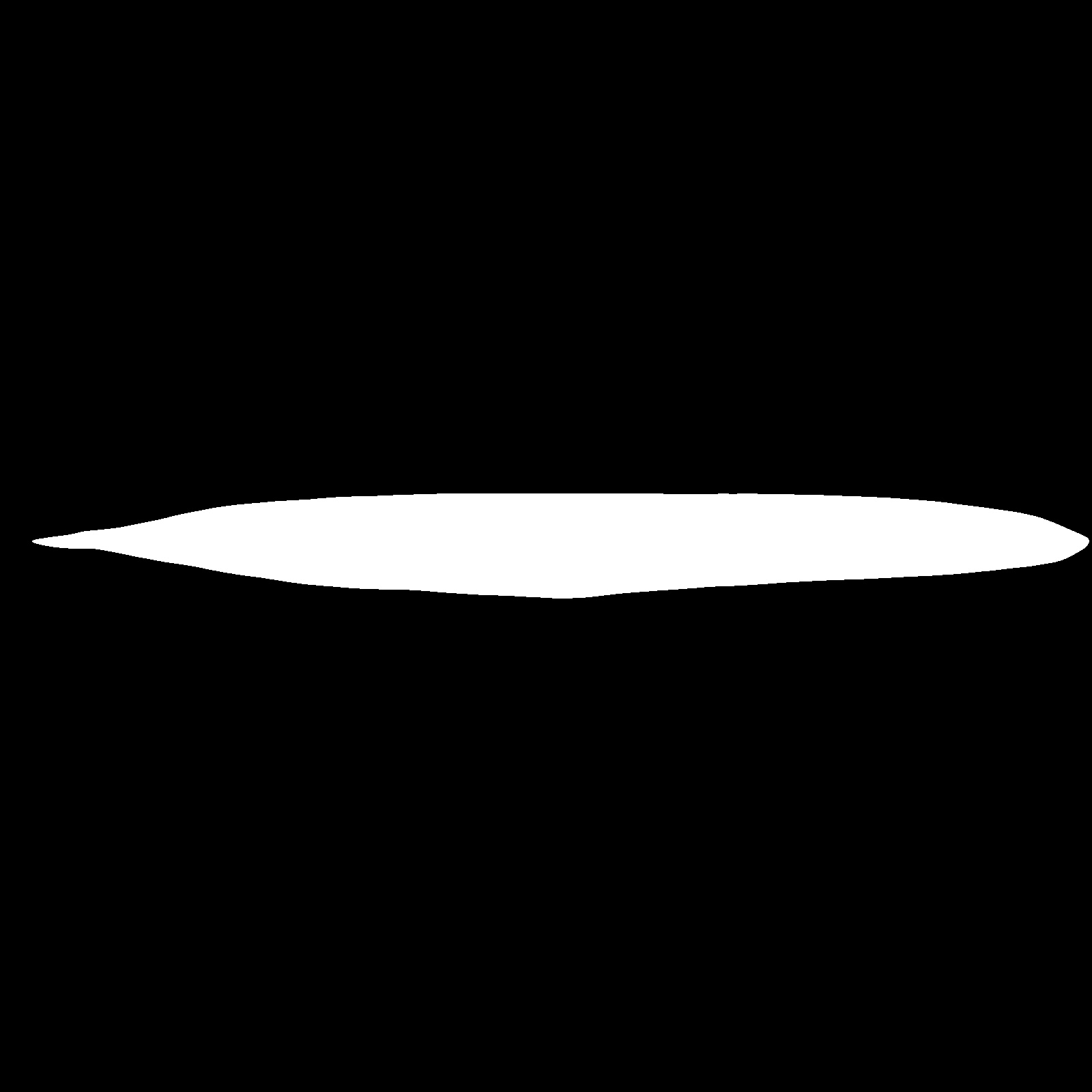}
       \caption{}
       \label{fig:f4}
    \end{subfigure}

    \caption{(a) Input Image (b) Grayscale Image (c) Binary Image }
    \label{fig: figure1}
    \centering
\end{figure}

After that, the binary images are typically obtained by thresholding the grey-level images. The algorithm used in the pre-processing phase is detailed in Figure \ref{fig: figure1}. The second column shows the image obtained after applying each step.

\subsubsection{Shape Feature Extraction}
\begin{figure}[!htpb]
    \centering
   \begin{subfigure}{0.3\textwidth}
       \centering
        \includegraphics[width=\linewidth]{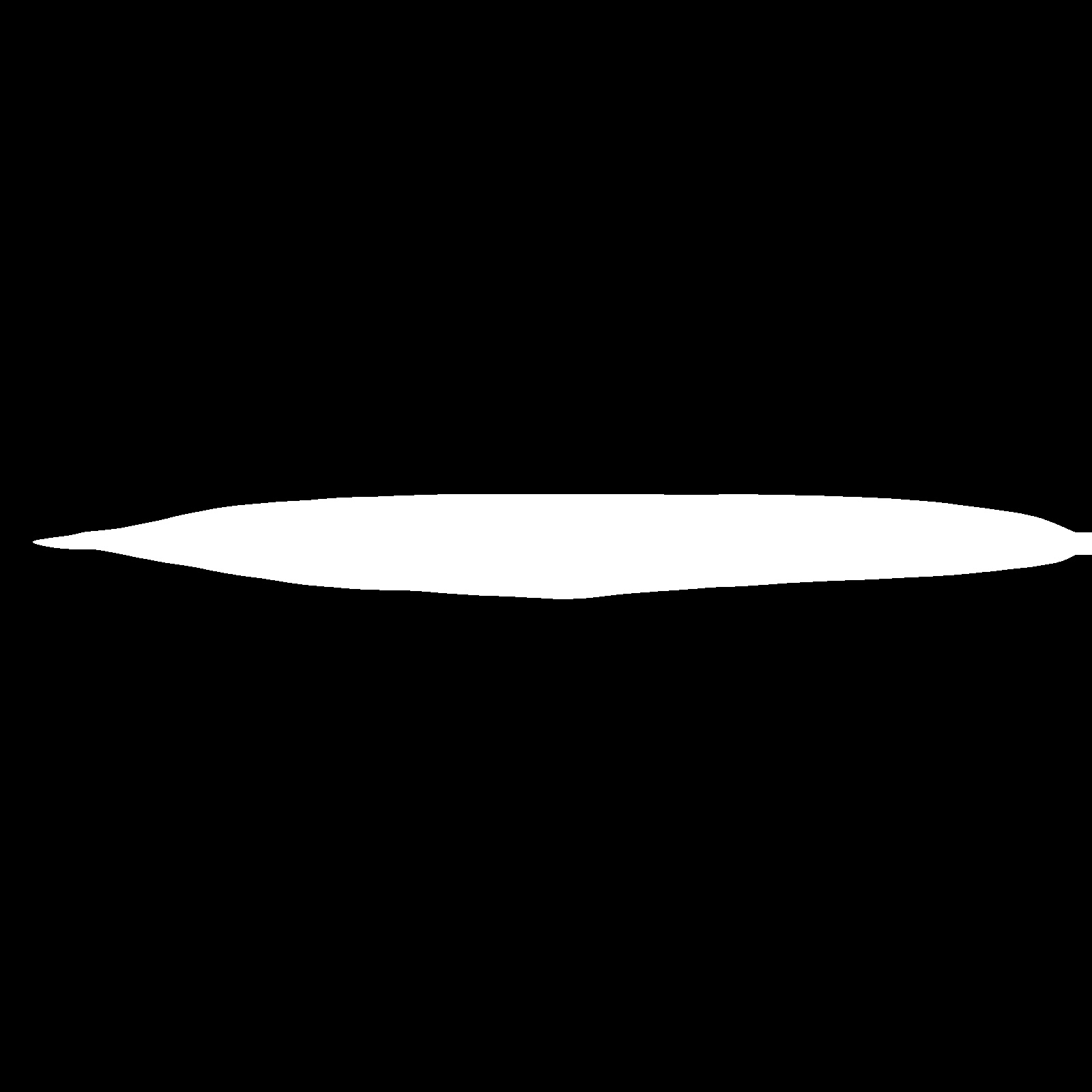}
       \caption{}
       \label{fig:f1} 
    \end{subfigure}\quad
    \begin{subfigure}{0.3\textwidth}
       \centering
        \includegraphics[width=\linewidth]{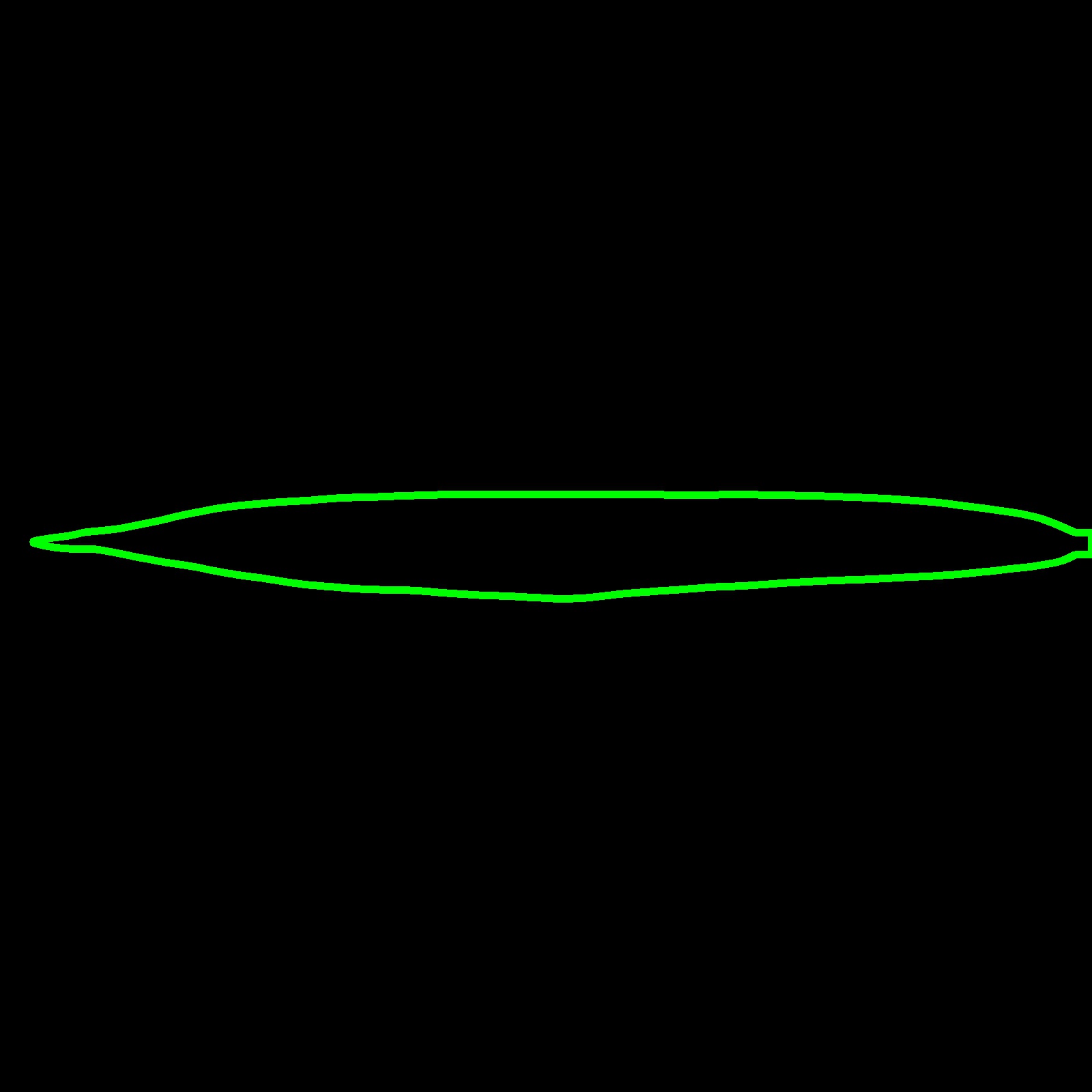}
       \caption{}
       \label{fig:f2}
    \end{subfigure}
    \caption{(a) Binary Image (b) Extracted Leaf Contour }
    \label{fig: figure2}
\end{figure}

Contour tracing is one of many preprocessing techniques performed on binary images to extract information about their general geometric features. Thus, it is necessary to find the contours of leaf images from the binary image set. The illustration is shown in Figure~\ref{fig: figure2}.\\
{\textbf{Geometric features}}

\begin{figure}[!htpb]
  \centering
  \includegraphics[width = \linewidth]{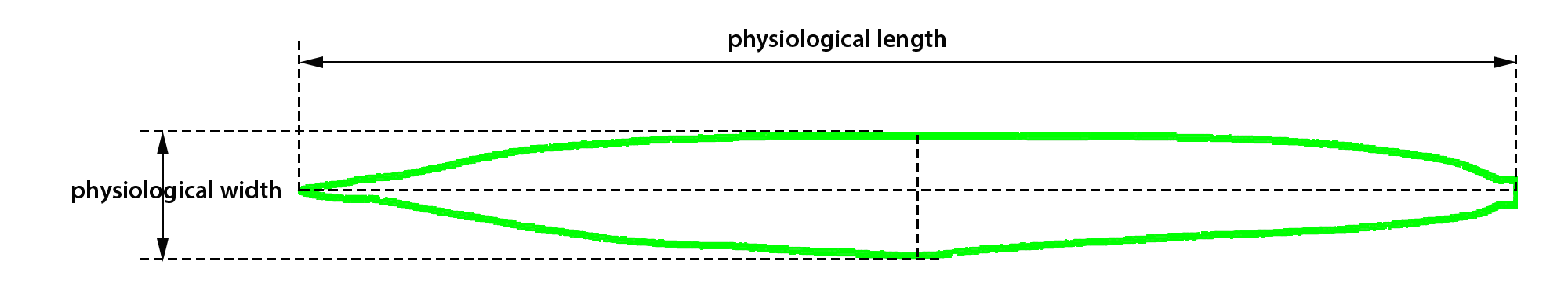}
  \caption{The relationship between physiological length and physiological width.} 
  \label{fig:figure3}
\end{figure} 

There are five basic geometric features: leaf length, leaf width, leaf area, and perimeter. \\
1) Leaf length\\
The leaf length is determined as the longest extension from the leaf apex to leaf base \cite{schrader2021leaf}. It can be denoted as $L_p$. \\
2) Leaf width \\
The leaf width lays perpendicular to the leaf axis length and corresponds to the longest distance. It is denoted as $W_p$. Figure~\ref{fig:figure3} shows the relationship between $L_p$ and $W_p$ in which the coordinates of two lines are orthogonal.\\
3) Leaf area \\
The value of leaf area is the total of pixels of binary value 1 on a smoothed leaf image. It is determined as $A$.\\
4) Leaf perimeter\\
The leaf perimeter is measured the length of the leaf contour, denoted as $P$.\\
5) Equivalent area diameter\\
Equivalent Area Diameter is defined as a diameter of a circle which has the same area as that of the leaf surface area, namely $D$, and can be computed as follows:
\begin{equation}
  D = \sqrt{\frac{4*A}{\pi}}
  \label{eqn: eq2}
\end{equation}
\begin{table}[!h]
   \caption{Five geometric attributes} 
   \label{tab:tab2}
   \centering 
   \begin{tabular}[t]{l>{\raggedright}p{0.1\linewidth}>{\raggedright\arraybackslash}p{0.3\linewidth}} 
   \textbf{Features} & \textbf{Symbol} & \textbf{Values (example)}\\ 
   \hline
   Physiological length & $L_p$ & 154 \\
   Physiological width & $W_p$ & 1552 \\
   Area & $A$ & 168711.5\\
   Perimeter & $P$ & 3248.9 \\
   Diameter & $D$ & 463.5 \\
   \hline
   \end{tabular}
\end{table}
For instance, one can extract the corresponding values of geometric features in a leaf image in Table \ref{tab:tab2}.\\
\textbf{Morphological features}\\
According to the research developed by S. G. Wu, \textit{et al., }\cite{wu2007leaf}, one can further extract six morphological features from calculated geometric features ($L_p, W_p, A, P$ and $D$).\\
1) Aspect ratio \\
It is the ratio of the physiological length to the physiological width. The formula is $L_p /W_p$.\\
2) Form factor\\
The form factor is used to describe the difference between a leaf and a circle. It can be computed using the formula: $4\pi A/P^2$.\\
3) Rectangularity\\
Rectangularity measures the similarity between a leaf shape and a rectangle, and is denoted as $L_pW_p/A$.\\
4) Narrow factor\\
It is a one of the main factor representing the surface characteristics, determined as the ratio of the leaf diameter and the leaf length, and is given by the formula $D/L_p$.\\
5) Perimeter ratio of diameter\\
This ratio describes the difference in size between leaf perimeter and leaf diameter. It is calculated by $P/D$.\\
6) Perimeter ratio of the leaf length and leaf width\\
Instead of measuring as its name, this parameter is measured by the ratio of leaf perimeter and the sum of the physiological dimensions, including length and width, thus $P/(L_p + W_p)$.
\begin{table}[!h]
   \caption{Six morphological features} 
   \label{tab:tab3}
   \centering 
   \begin{tabular}[t]{p{0.3\linewidth}>{\raggedright}p{0.15\linewidth}>{\raggedright\arraybackslash}p{0.25\linewidth}} 
   \textbf{Features} & \textbf{Symbol} & \textbf{Values (example)}\\ 
   \hline
   Aspect ratio & $L_p /W_p$ & 10.08 \\
   Form factor &  $4\pi A/P^2$ & 0.20 \\
   Rectangularity & $L_pW_p/A$ & 1.42\\
   Narrow factor & $D/L_p$ & 3.01 \\
   Perimeter ratio of diameter & $P/D$ & 7.01 \\
   Perimeter ratio of the leaf length and leaf width & $P/(L_p + W_p)$ & 1.90\\
   \hline
   \end{tabular}
\end{table}
Taking a random leaf image in the Flavia dataset as an example, one can extract six values of morphological features, respectively in Table~\ref{tab:tab3}.\\
\textbf{Moment features}\\
For image processing in general and computer vision in particular, an image moment illustrates the image pixel intensities are covered in terms of its distribution in its orientation. In other words, it is defined as a certain particular weighted average of image pixels' intensities. There are three different types of moments: spatial moments, central moments, and central normalized moments, including 24 various moment features.\\
1) Spatial moments\\
For each image $I$, one can compute the spatial moments as follows:
\begin{equation*}
 M_{i,j} = \sum_x\sum_yx^iy^jI(x,y).
\end{equation*}
In this paper, we select the following spatial moments from a given leaf image: $M_{0,0}$, $M_{0,1}$, $M_{0,2}$, $M_{0,3}$, $M_{1,0}$, $M_{1,1}$, $M_{1,2}$, $M_{2,0}$, $M_{2,1}$, and $M_{3,0}$.\\
2) Central moments\\
The central moments can be calculated as:
\begin{equation*}
 \mu_{p,q} = \sum_x\sum_y(x - \Bar{x})^p(y - \Bar{y})^qI(x,y),
\end{equation*}
where $p$ and $q$ are adjusted. There are seven parameters to be attained, consists of $\mu_{0,2}$, $\mu_{0,3}$, $\mu_{1,1}$, $\mu_{1,2}$, $\mu_{2,0}$, $\mu_{2,1}$, and $\mu_{3,0}$, 
The centroid point can be computed as:
\begin{equation*}
 \Bar{x} = \frac{M_{1,0}}{M_{0,0}} \hspace{2em} \Bar{y} = \frac{M_{0,1}}{M_{0,0}}
\end{equation*}
3) Normalized central moments\\
The normalized central moments can be extracted as
\begin{equation*}
 \eta_{i,j} = \frac{\mu_{i,j}}{\mu_{0,0}^{1 + \frac{i + j}{2}}}
\end{equation*}
In our work, we select the following normalized central moments: $\eta_{0,2}$, $\eta_{0,3}$, $\eta_{1,1}$, $\eta_{1,2}$, $\eta_{2,0}$, $\eta_{2,1}$, and $\eta_{3,0}$.

\subsubsection{Texture Feature Extraction}
Many studies have applied texture features in pattern recognition and image processing due to their essential information. In this research, texture features are calculated from a gray level co-occurrence matrix (GLCM). Based on the second moment, fourteen texture features describe the relationship among pixels in images \cite{benvco2007novel}. With regards to the distance $d$, GLCM is computed from the joint probabilities between pairs of pixels, which are two gray levels i, j at a given offset according to a given direction $\theta$ \cite{haralick1973textural}.
In terms of the scale-invariant of the texture pattern, the GLCM is standardized by total pairs of pixels in the following equation:
\begin{equation}
 p_{d,\theta}(i,j) = \frac{P_{d,\theta}(i,j)}{N_p},
 \label{eqn: eq3}
\end{equation}
where $N_p$ is the total number of pairs of pixels in the image. 

Noticeably, GLCM was proposed by Haralick \cite{haralick1973textural} for texture descriptions in the 1973s and has become one of the most well-known and widely used texture measures. Haralick proposed fourteen measures (as defined in Table ~\ref{tab:tab5}) of textural features derived from the co-occurrence matrix. For the sake of ease, we will use the notation $ p(i,j)$ instead of using $ p_{d,\theta}(i,j)$ in the rest of the paper.
\begin{table}[!http]
	\centering
	\caption{Fourteen Haralick Features}
	\label{tab:tab5}
	\begin{adjustbox}{}
		\begin{tabular}{|p{5cm}| p{5cm}|}
			\hline
			Feature & Mathematical Expression \\    
			\hline 
			Angular Second Moment & $ f_{1} = \sum_{i}\sum_{j}{p(i,j)}^2$\\
			\hline
			Contrast & $ f_{2} = \sum_{i,j}|i - j|^2 p(i,j) $ \\
			\hline
			Correlation & $f_{3} = \frac{\sum_{i}\sum_{j}(ij)p(i,j) - \mu_x\mu_y}{\sigma_x\sigma_y}$\\
			\hline
			Sum of Squares: Variance & $f_{4} = \sum_{i}\sum_{j} (i-\mu)^2p(i,j)$\\
			\hline
			Inverse Difference Moment & $f_{5} = \sum_{i}\sum_{j}\frac{p(i,j)}{1+ (i - j)^2}$\\
			\hline
			Sum Average & $f_{6} = \sum_{i = 2}^{2N}ip_{x+y}(i)$\\
			\hline
			Sum Entropy & $f_{8} = -\sum_{i = 2}^{2N}p_{x+y}(i)\log\{p_{x+y}(i)\}$\\
			\hline
			Entropy & $f_{9} = -\sum_{i}\sum_{j}p(i,j)\log(p(i,j))$\\
			\hline
			Difference Variance & $f_{10} = -\sum_{i = 0}^{N - 1}i^2p_{x - y}(i)$\\
			\hline
			Difference Entropy & $f_{11} = -\sum_{i = 0}^{N - 1}p_{x - y}(i)\log\{p_{x-y}(i)\}$\\
			\hline
			Infomation measure of correlation 1 & $f_{12} = \frac{HXY - HXY1}{\max\{HX,HY\}}$\\
			\hline
			Infomation measure of correlation 2 & $f_{13} = (1 - \exp[-2(HXY2- HXY)])^\frac{1}{2}$\\
			\hline
			Maximal Correlation Coefficient & $f_{14} = \left({\lambda_2(Q)}\right)^\frac{1}{2}$, where $\lambda_2$ is the second largest eigenvalue of the matrix $Q$.\\
			\hline
		\end{tabular}
	\end{adjustbox}
\end{table}
In Table \ref{tab:tab5}, $p_x$ and $p_y$ are represented by two statistical measures - mean and standard deviation - consisting of $\mu_x$, $\mu_y$, $\sigma_x$, and $\sigma_y$. Moreover, according to the coordinates (row $x$ and columns $y$) of an entry in the co-occurrence matrix, the value $p_{x\pm y}(i)$ is the probability of co-occurrence matrix coordinates summing to $x\pm y$. Also, $HXY = f_9$, $HX$ and $HY$ are entropies of $p_x$ and $p_y$ where:
\begin{eqnarray}
HXY1 &=& -\sum_{i}\sum_{j}p(i,j\log(p_x(i)p_y(j)), \\
HXY2 &=& -\sum_{i}\sum_{j}p_x(i)p_y(j)\log(p_x(i)p_y(j)),\\
Q(i,j) &=& \sum_k \frac{p(i,k)p(j,k)}{p_x(i)p_y(k)},\\
p_x(i) &=& \sum_{j=1}^N p(i,j), \\
p_y(j) &=& \sum_{i=1}^N p(i,j).
\end{eqnarray}

\subsubsection{Color Feature Extraction}
\begin{figure}[!h]
    \centering
   \begin{subfigure}{0.4\textwidth}
       \centering
        \includegraphics[width=\linewidth]{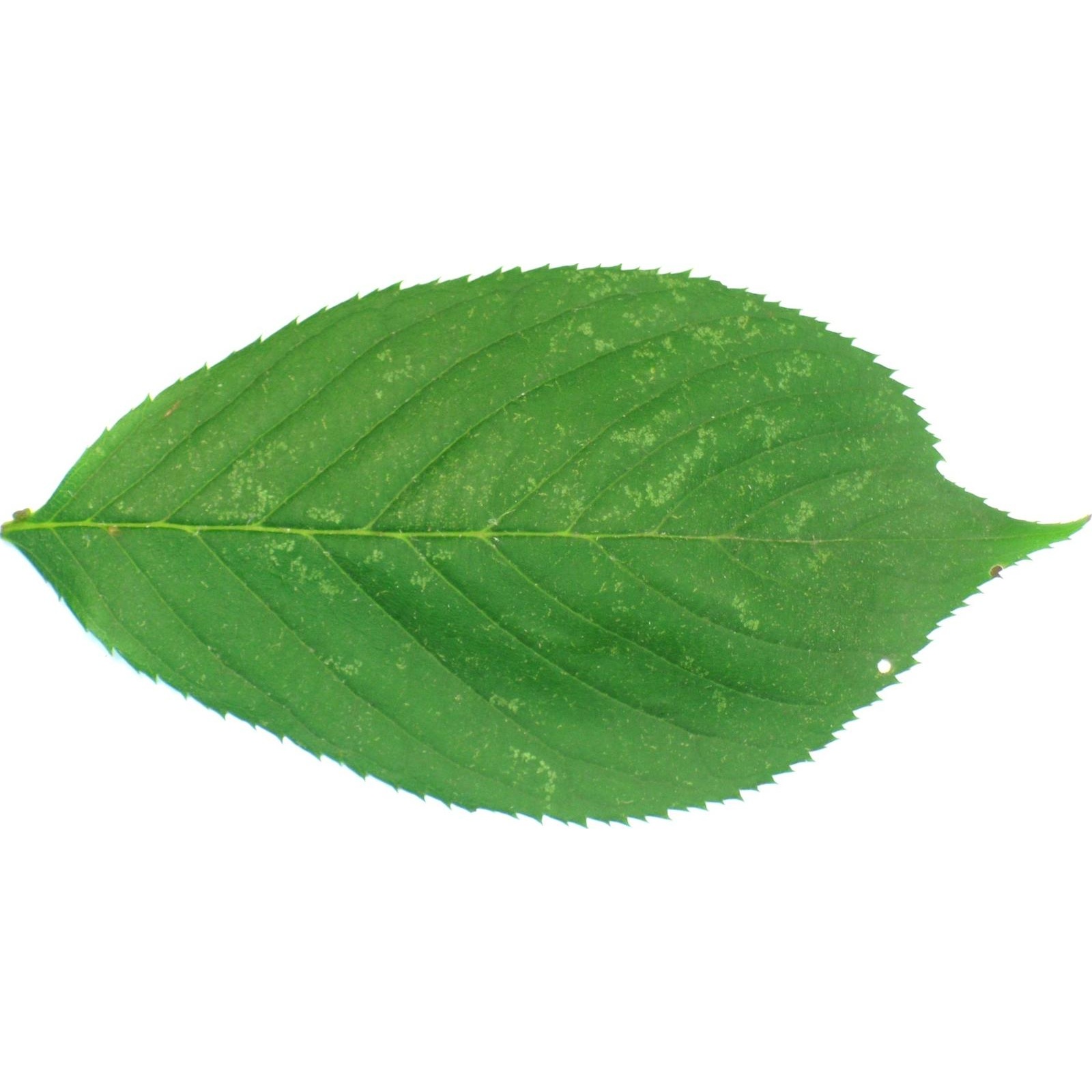}
       \caption{}
       \label{fig:f1} 
    \end{subfigure}
    \begin{subfigure}{0.4\textwidth}
       \centering
        \includegraphics[width=\linewidth]{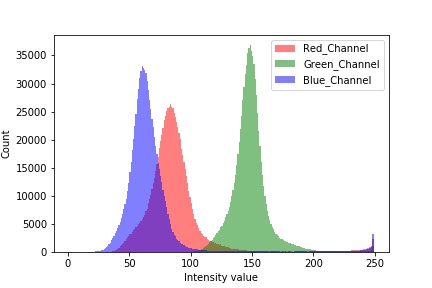}
       \caption{}
       \label{fig:f2}
    \end{subfigure}\\
    \begin{subfigure}{0.4\textwidth}
       \centering
        \includegraphics[width=\linewidth]{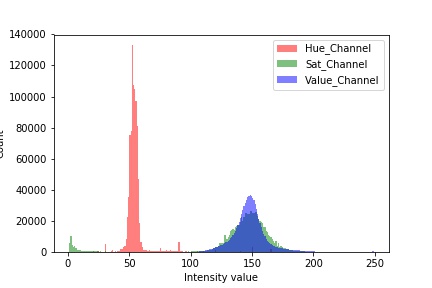}
       \caption{}
       \label{fig:f3}
    \end{subfigure}
    \begin{subfigure}{0.4\textwidth}
       \centering
        \includegraphics[width=\linewidth]{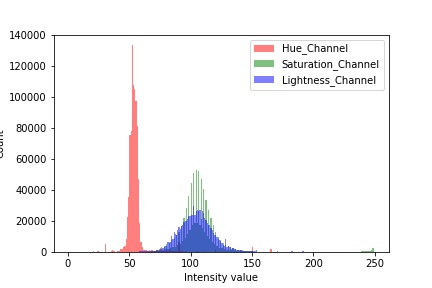}
       \caption{}
       \label{fig:f3}
    \end{subfigure}
    \caption{(a) Leaf sample (b) RGB histogram (c) HSV histogram (d) HSL histogram} 
    \label{fig: figure6}
    \centering
\end{figure}

The color feature is one of the most commonly used attributes in image procession. Indeed, colors can contain many features based on selected color space. In this paper, RGB, HSV, and HSL are three chosen spaces. The feature includes four descriptive statistics (mean, variance, skewness, and kurtosis) for an image that requires the image's histogram. One can determine the formula for each statistical measurement in Table~\ref{tab:tab4}.\\
\textbf{RGB color space} \\
Normally, RGB is taken priority over others due to its representative. Red, green, and blue are three components of this color space. The color subspace of interest is organized in a unit cube with coordinates $(R, G, B)$, where the values are more spread out from black $(0, 0, 0)$ to white $(1, 1, 1)$.\\
\textbf{HSV color space}\\
The HSV model was born in 1978 and invented by Alvy Ray Smith. In this model, the ``H''denoted the Hue, ``S'' is short for the Saturation, and ``V'' is the value. HSV is often utilized in computer vision and image analysis for the segmentation process.\\
\textbf{HSI color space}\\
Similar to HSV, in HSI space, ``H'' and ``S'' referred to the Hue and Saturation, but ``I'' indicates the Intensity \cite{wen2004color}. Another name of this model is HSL, where ``L'' is short for lightness. One can visualize the leaf sample in three color spaces in Figure~\ref{fig: figure6}.
\begin{table}[!h]
\centering
\caption{Color features}
\label{tab:tab4}
\begin{adjustbox}{}
    \begin{tabular}{|p{4cm}| p{5cm}|}
    \hline
    Parameter & Mathematical Equation \\    
    \hline 
    Mean & $\mu = \frac{1}{N}.\sum_{i = 1}^N(x_i)$\\
    \hline
    Variance & $\sigma^2=\frac{1}{N}.{\sum_{i=1}^N(x_i-\mu)^2}$ \\
    \hline
    Skewness & $\gamma_1 = \frac{1}{N}.{\sum_{i=1}^N.[\frac{x_i - \mu}{\sigma}]^3}$\\
    \hline
    Kurtosis & $\gamma_2 = \frac{1}{N}.{\sum_{i=1}^N.[\frac{x_i - \mu}{\sigma}]^4} - 3$\\
    \hline
    \end{tabular}
\end{adjustbox}
\end{table}

\subsubsection{Fourier Descriptor Extraction}
The Fourier Descriptor method was first introduced by Zahn and Roskies in 1972 \cite{zahn1972fourier}. To implement this method, we extract the boundary points of the leaf region after converting them into the binary image.
Assume that BP denotes the boundary pixel set and C denotes the number of boundary pixels, then the centroid of the object $(x_c, y_c)$ can be represented as
\begin{eqnarray}
  x_c = \frac{1}{C}{\sum_{t = 1}^Cx(t)},
  y_c = \frac{1}{C}{\sum_{t = 1}^Cy(t)},
\end{eqnarray}
where $(x(t),y(t))$ are  the corresponding location of the t-th pixel of
the set BP in a gray image.

Assuming $r(t)$ as the radius of the t-th pixel of the set BP to the centroid; then, it can be formulated as:
\begin{equation}
  r(t) = \sqrt{(x(t) - x_c)^2 + (y(t) - y_c)^2} \hspace{1em}t = 1,2..,C
\end{equation}
After that, the Fast Fourier Transformation (FFT) is applied on each of the component $x$, $y$ respectively to form the Fourier Descriptor feature vector \cite{turkoglu2019recognition}.
\begin{equation}
  F(i) = FFT\{r\}_i
\end{equation}

\subsubsection{Vertical and Horizontal Projection}

\begin{figure}[!h]
	\centering
	\begin{subfigure}[b]{0.45\linewidth}
		\includegraphics[width=\linewidth]{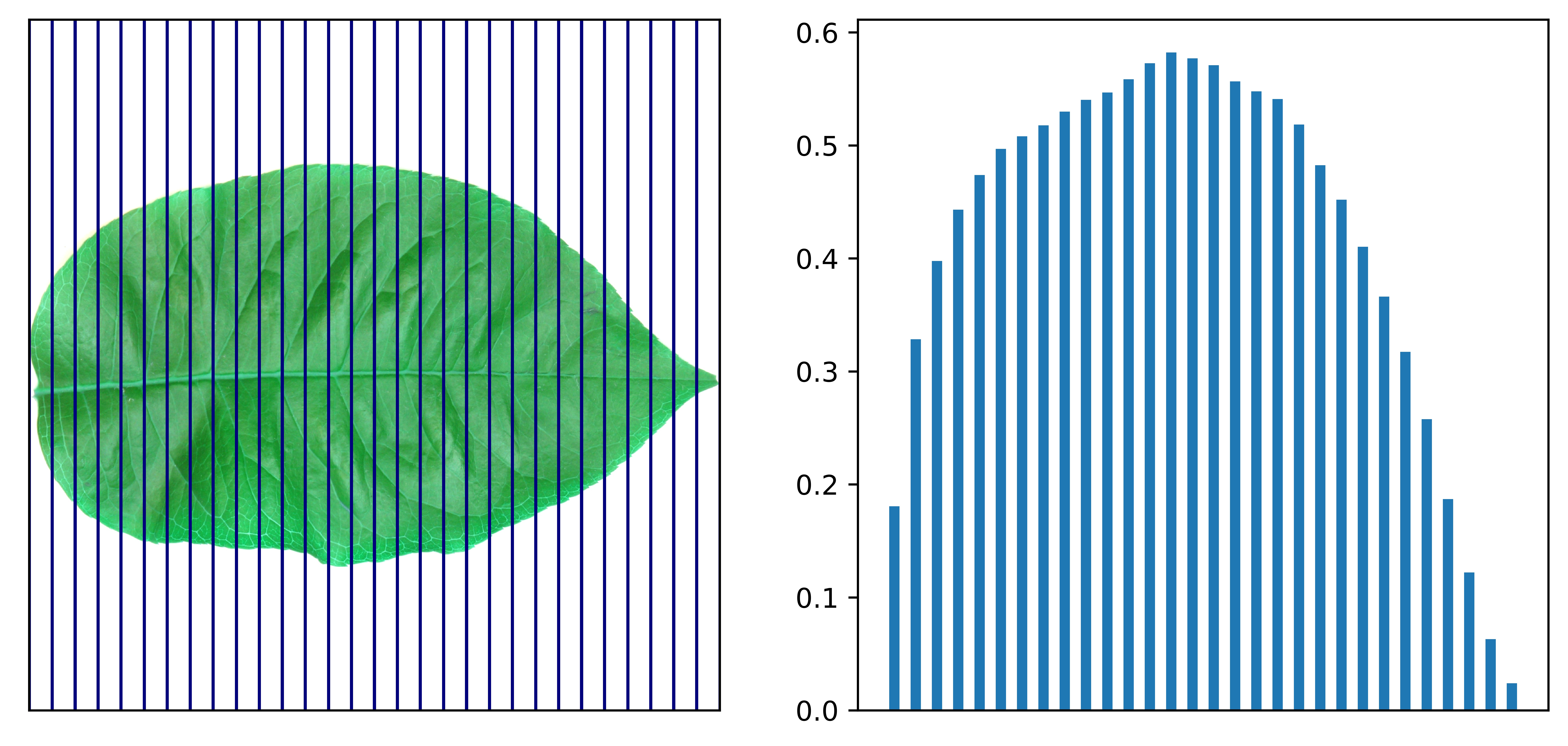}
		\caption{Vertical projection histogram}
		\label{fig:yprojection}
	\end{subfigure}
	\begin{subfigure}[b]{0.45\linewidth}
		\includegraphics[width=\linewidth]{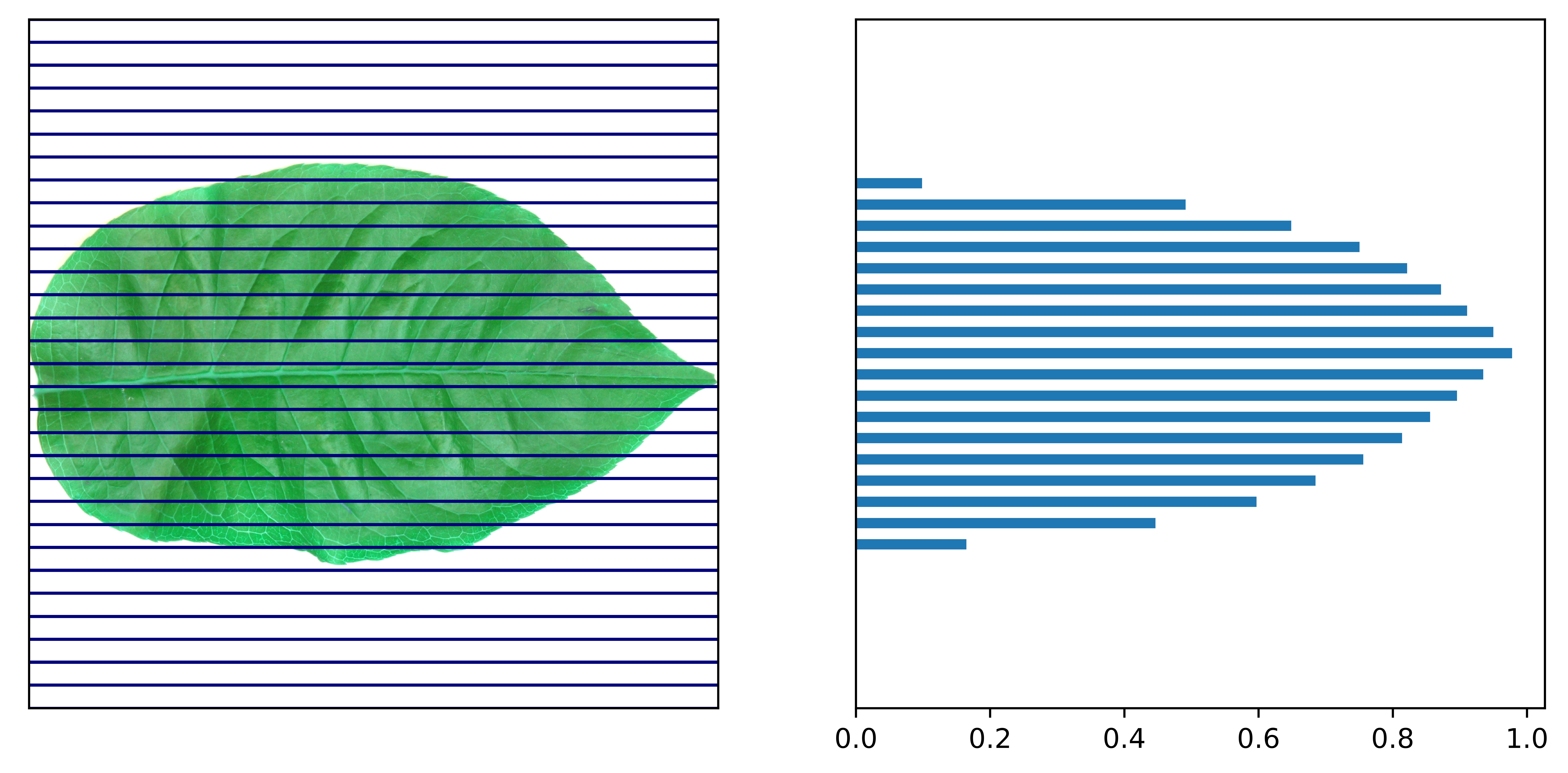}
		\caption{Horizontal projection histogram}
		\label{fig:xprojection} 
	\end{subfigure}
	\caption{Virtical and horizontal projeciton}
	\label{fig:xyprojection}
	\centering
\end{figure}

This feature primarily represents the shape of rotated leaf images concerning $x$ and $y$ axes. The mask that indicates which pixels belong to the rotated leaf is projected onto two sides of the bounding square, producing two distributions of the leaf pixels along the axes. In specific, the rotated, cropped image is split into 30 bins along each axis. The percentage of leaf pixels out of all pixels is then computed in every bin. Hence in total, we have a vector of length 60 consisting of two so-called xy-projection histograms. Figure \ref{fig:xyprojection} demonstrates how xy-projection histograms are calculated.

\subsubsection{Vein feature extraction}
Vein images can be obtained by applying morphological operations such as opening, closing, dilation, and erosion. The processing is illustrated as following: \\
{\textbf{Step 1}}. Apply Gaussian Filter on each image 1600 x 1600 with the kernel (25,25). \\
{\textbf{Step 2}}. Create disk-shaped structuring elements of radius 1,2,3, and 4 respectively on the grayscale images. \\
{\textbf{Step 3}}. These structural elements are performed by erosion and dilation. \\
{\textbf{Step 4}}. The operation creates subtracted grayscale images of the leaf. \\
The vein leaves will be extracted vein features by applying the CNN model.

\subsection{Our proposed classification model}

In this section, we describe in detail our proposed model for leaf classification. The model consists of multiple encoders and a single decoder. After doing the image processing, the processed color image, vein image, and handcrafted features are fed into the encoders. Each encoder transforms the input into its best-learned representation. Outputs from decoders are then concatenated into a single vector, on which the decoder will make the final prediction. 

\subsubsection{Architecture}

Given the fact that the model utilizes information from various sources - color, vein images, and other handcrafted features, the encoders and decoder do not need to be very large and complicated. Encoders are deliberately designed to exploit its input and project the information into 100-dimensional spaces. The consistent length of encoders' outputs makes none of the feature groups dominate the final prediction. We choose 100 dimensions after some experiments seeing that it is reasonable to compress information from color images. The most suitable architectures for encoders are probably neural networks to learn the best representations of the information. 

For images, CNNs have been proved to be the most effective models to capture spatial features \cite{yamashita2018convolutional}. Therefore, we design those encoders as some stacked convolutional layers followed by a fully connected layer. In contrast, handcrafted features are manually extracted through image processing and ready-to-use; thus, we only use a single fully connected layer to encode these features. On the other hand, due to our training settings, the topology of output space from decoders is quite simple and can be easily captured by a decoder such as a support vector machine.

\begin{figure}[h!]
	\centering
	\includegraphics[width=1.0\linewidth]{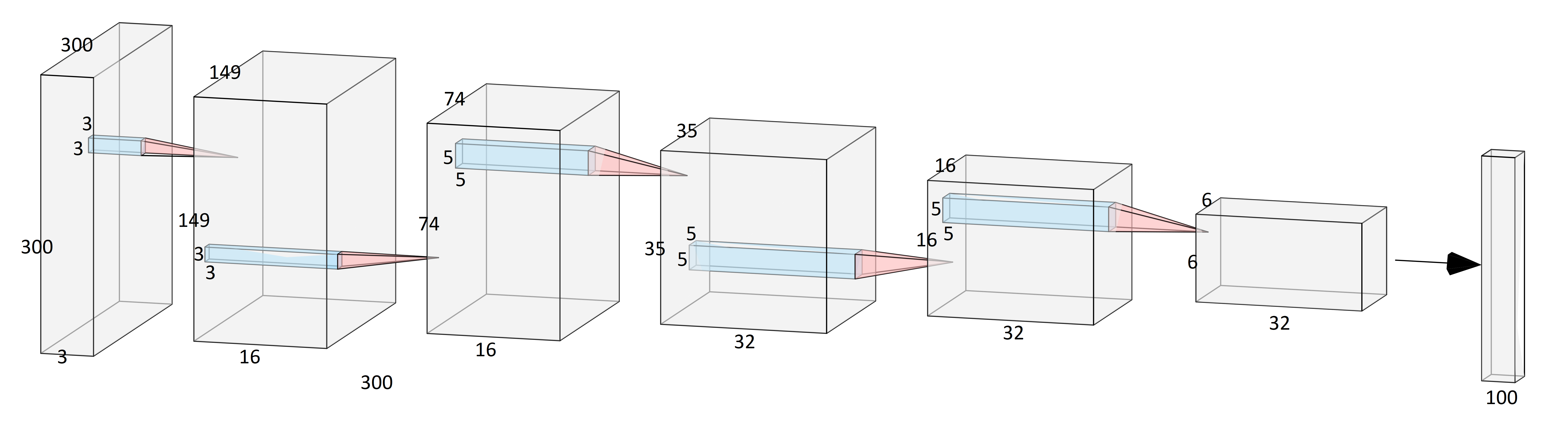}
	\caption{Our proposed 2-dimensional convolutional encoder}
	\label{fig:CNN_component}
\end{figure}

In total, there are seven encoders: two 2-dimensional CNNs for color and vein, one 1-dimensional CNN for xy-projection histogram, and four fully-connected networks for handcrafted features. 
Firstly, CNN encoders, with the role of extracting spatial features from processed color images or vein images, consist of 5 two-dimensional convolutional layers. The first two layers have 16 filters of size $3 \times 3$, and the following three layers have 32 filters of size $5 \times 5$. They all use Rectified Linear Unit (ReLU) as the activation function and only compute at valid pixels without any padding. Each of the convolutional layers is followed by a batch normalization layer and a max-pool layer. At the end of the encoder, the image is flattened and input to a fully connected layer of 100 units. Figure \ref{fig:CNN_component} depicts the CNN encoders as described.

Secondly, xy-projection histogram has 1-dimensional spatial features and is likely to be well captured by 1-dimensional CNN. Similar to 2-dimensional encoders, we use three blocks of convolutional layer, batch normalization, and max-pooling. The first two layers have 16 filters, and the other uses 32 filters. They all use $3 \times 3$ filter size, ReLU activation function, and compute only at valid windows. 

Thirdly, encoders that deal with handcrafted features are simple because these features already contain ready-to-use information. Each branch has only a fully connected layer, transforming its features into a 100-dimensional space. Finally, we use the activation function ReLU to create non-linearity.

Lastly, to make a prediction, extracted information from encoders is combined by concatenating the 100-dimensional vectors into a single 700-dimensional vector. As the encoders are separately trained, the vector should be normalized to a mean of 0.0 and standard deviation of 1.0 before feeding into SVM to decide which class the leaf belongs to.

\subsubsection{Training}
\label{subsec:model:training}
All the encoders and decoders are trained separately so that each encoder can learn the best representation for the input images. 
One softmax layer is added on top of the encoder, then both of them are jointly trained to classify which class the input is from for training an encoder. 

We aim to classify leaves into the corresponding categories; therefore, training encoders for the current leaf classification problem makes them learn the representations towards the task. Moreover, this adding a dense layer also makes the output space easy to exploit by an SVM classifier with kernel Radial Basis Function as the classifier. A grid search is performed on validation sets to look for the best gamma parameter and penalty parameter $C$. Besides, to prevent overfitting, L2 regularization and dropout are applied during training.

\section{Experiments and Results}

In this section, we first explain how we assess our models by 10-fold cross-validation. Then we report how much each feature group can contribute to the classification problem and the final performance of the entire model.

\subsection{Experimental settings}


To validate our method, we conduct 10-fold cross-validation in which the dataset is split into ten complementary folds. We use one fold for testing in each round, and the others are utilized for training and validation. In specific, images are sorted by their original filenames from the link\footnote{http://flavia.sourceforge.net/}, and consecutively assigned to be on the test set or validation set. This split is inspired by \cite{turkoglu2019recognition} where they also use 10-fold cross-validation on the Flavia dataset. It is a little different than our last seven images still follow the procedure of assigning. Table \ref{table:kfolds_indexed} shows how the first 10 images are distributed into 10 groups. 

\begin{table}[h!]
	\centering
	\scalebox{0.81}{\begin{tabular}{||l||c|c|c|c|c|c|c|c|c|c|}
			\hline \hline
			Filename & Fold\_1 & Fold\_2 & Fold\_3 & Fold\_4 & Fold\_5 & Fold\_6 & Fold\_7 & Fold\_8 & Fold\_9 & Fold\_10 \\ \hline \hline
			1001.jpg & \cellcolor{blue!25}Test    &  \cellcolor{blue!10}Valid   & Train   & Train   & Train   & Train   & Train   & Train   & Train   & Train    \\ \hline
			1002.jpg & Train   & \cellcolor{blue!25} Test    & \cellcolor{blue!10}Valid   & Train   & Train   & Train   & Train   & Train   & Train   & Train    \\ \hline
			1003.jpg & Train   & Train   & \cellcolor{blue!25}Test    & \cellcolor{blue!10}Valid   & Train   & Train   & Train   & Train   & Train   & Train    \\ \hline
			1004.jpg & Train   & Train   & Train  & \cellcolor{blue!25}Test    & \cellcolor{blue!10}Valid   & Train   & Train   & Train   & Train   & Train    \\ \hline
			1005.jpg & Train   & Train   & Train   & Train   & \cellcolor{blue!25}Test    & \cellcolor{blue!10}Valid   & Train   & Train   & Train   & Train    \\ \hline
			1006.jpg & Train   & Train   & Train   & Train   & Train   & \cellcolor{blue!25}Test    & \cellcolor{blue!10}Valid   & Train   & Train   & Train    \\ \hline
			1007.jpg & Train   & Train   & Train   & Train   & Train   & Train   & \cellcolor{blue!25}Test    & \cellcolor{blue!10}Valid   & Train   & Train    \\ \hline
			1008.jpg & Train   & Train   & Train   & Train   & Train   & Train   & Train   & \cellcolor{blue!25}Test    & \cellcolor{blue!10}Valid   & Train    \\ \hline
			1009.jpg & Train   & Train   & Train   & Train   & Train   & Train   & Train   & Train   & \cellcolor{blue!25}Test    & \cellcolor{blue!10}Valid    \\ \hline
			1010.jpg & \cellcolor{blue!10}Valid   & Train   & Train   & Train   & Train   & Train   & Train   & Train   & Train   & \cellcolor{blue!25}Test     \\ \hline
	\end{tabular}}
	\caption{Indexed 10-fold cross validation}
	\label{table:kfolds_indexed}
\end{table}

During the training session for each fold, the training set is used as actual training samples. The validation set is for selecting the best parameters and hyper-parameters of the models. In contrast, the test set is responsible for evaluating the performance after training. 

The mean and standard deviation of accuracy is computed over ten folds are the performance metric. Finally, in addition to the above particular partition, we perform random 10-fold cross-validation to conclude our method's final performance. In that, the folds are randomly split but preserve the percentage of samples for each class.

\subsection{Individual feature groups}
\label{subsec:individual_group}

\begin{figure}[h!]
	\includegraphics[width=\linewidth]{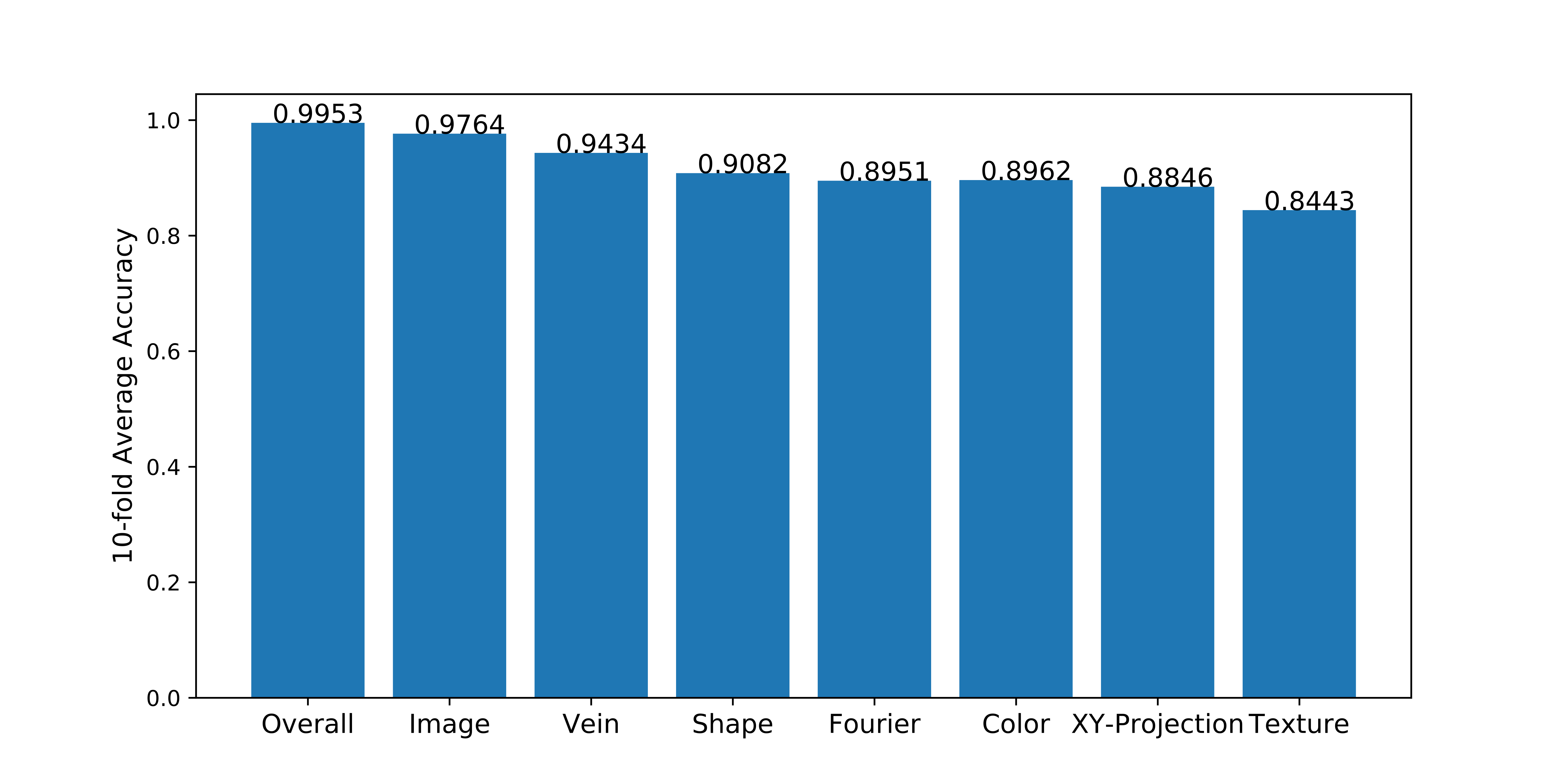}
	\caption{Accuracy of feature groups over 10 folds.}
	\label{fig:individual_performance}
\end{figure}

As discussed in Section \ref{subsec:model:training}, the encoders have the described architectures but adding a single softmax layer to train as a classifier. We tune $L_2$ regularization rates and dropout rates to have the models give the best results.
Figure \ref{fig:individual_performance} plots the test accuracy of each feature group under 10 fold cross-validation. Generally, the accuracy of each modality was larger than 80\%. Two techniques using vein features and shape features were higher classification performance with 94.34\% and 90.82\%, respectively. Although according to the analysis, the result utilizing color images achieved the highest accuracy compared with others. However, this modality contained overlapped features from other subsets. 

Indeed, the raw images had to conclude all features related to shape, texture, color, vein attributes since we also extracted these kinds of features from their origin. Nevertheless, one may not determine which features belong to each modality due to CNN's technique. Therefore, in terms of individual modality, color images would not be taken into account. By considering the color features, the accuracy obtained was 89.62\%, followed by Fourier descriptors with 89.51\%. XY-projector features could be regarded as histogram features and provided an accuracy of 88.46\%. Finally, texture attributes ended with the lowest rank at 84.43\%.

Looking at the details, the performances of both shape and Fourier features were nearly the same. Therefore, one of the most reasonable explanations could be that the Fourier descriptors were extracted based on the contour of the leaf, which represented most characteristics of a leaf's shape. However, as the number of measures of these features was unequal, this led to differences in their accuracy. 

Similar to the pair of shape and Fourier features, the performances of color and xy-projection showed a slight difference, around 1\%. One of the possible root causes is the histogram of leaf pictures was counted according to the frequency of luminance in the image while calculating the illumination based on the values of three color channels. However, the accuracy of color modality was still higher since we used many kinds of color models which had more useful features in classification tasks.  

In order to see the full performance of the proposed method, we combined all modalities and evaluated again with random 10-fold cross-validation. Our best model was found at $99.79\% \pm 0.35\% $ on validation sets and $99.53\% \pm 0.37\% $ on test sets. The performances were almost the same, $99.74\% \pm 0.42\%$ on validation sets and $99.58\% \pm 0.31\%$ on test sets. Until now, the best recognition rate for the Flavia dataset was  99.43\% \cite{su2020fast}. Thus, our experimental results show that the proposed approach could bypass the current state-of-the-art methods. 

\section{Conclusion and future work}

We have proposed an approach for leaf classification which yields a nearly perfect result, $99.58\%\pm 0.32\% $ on test sets under random 10-fold cross-validation. The proposed methodology is not limited to apply for leaf classification tasks. Still, it is also applicable to other works that lack data or computation resources and need handcrafted features to improve the classification performance.

However, our encoders are theoretically not optimal. They are separately trained and most likely to capture overlapped information. For example, one could withdraw any feature groups apart from color image such as vein image, color, shape, texture from a color image. The CNN encoder may extract the same features as other encoders since they do not know each other. But in contrast, this setting is beneficial to the flexibility of encoders, and they can capture the best representations of the feature groups.

On the other hand, our study mainly focuses on leaves recorded under an ideal laboratory condition. The images are assumed to be already processed to remove background and noise. This refined version is far from realistic and partially explains the extremely high performance. In the future work, we will apply other different approaches for feature extraction and utilize some recent deep learning techniques to enhance the performance of the main problem.

\section*{Acknowledgments}
This research is funded by Vietnam National University Ho Chi Minh City (VNU-HCM) under grant number NCM2019-18-01. We want to thank the University of Science, Vietnam National University in Ho Chi Minh City, and  AISIA Research Lab in Vietnam for supporting us throughout this paper.

\bibliographystyle{spbasic}      
\bibliography{references}   

\end{document}